\documentclass[11pt]{article}
\usepackage[utf8]{inputenc}
\usepackage{amsmath}
\usepackage{graphicx}
\usepackage[numbers]{natbib}
\usepackage{booktabs}
\usepackage{geometry}
\usepackage{adjustbox}
\usepackage{url}
\title{Predictive Digital Twins for Thermal Management Using Machine Learning and Reduced-Order Models}
\author{Tamilselvan Subramani\thanks{HELLA India Automotive Pvt. Ltd., Chennai, TN, India. Email: thamizhselvan.subramani@gmail.com} \and Sebasti\'{a}n Bartscher\thanks{HELLA GmbH \& Co. KGaA, Lippstadt, Germany. Email: sebastian.bartscher@forvia.com}}
\date{May 2025}

\begin{document}

\maketitle

\begin{abstract}
Digital twins enable real-time simulation and prediction in engineering systems. This paper presents a novel framework for predictive digital twins of a headlamp heatsink, integrating physics-based reduced-order models (ROMs) from computational fluid dynamics (CFD) with supervised machine learning. A component-based ROM library, derived via proper orthogonal decomposition (POD), captures thermal dynamics efficiently. Machine learning models, including Decision Trees, k-Nearest Neighbors, Support Vector Regression (SVR), and Neural Networks, predict optimal ROM configurations, enabling rapid digital twin updates. The Neural Network achieves a mean absolute error (MAE) of 54.240, outperforming other models. Quantitative comparisons of predicted and original values demonstrate high accuracy. This scalable, interpretable framework advances thermal management in automotive systems, supporting robust design and predictive maintenance.

\end{abstract}

\section{Introduction}
Digital twins, virtual replicas of physical systems, are revolutionizing automotive engineering by enabling real-time monitoring and optimization. Thermal management of components like headlamp heatsinks is critical for performance and durability. High-fidelity computational fluid dynamics (CFD) simulations are accurate but computationally expensive, limiting their use in real-time applications. Reduced-order models (ROMs) reduce computational cost, but their application to complex, nonlinear systems remains challenging.

This paper, adapted from \cite{subramani2022}, proposes a framework for predictive digital twins using component-based ROMs and supervised machine learning. Proper orthogonal decomposition (POD) creates a scalable ROM library from CFD simulations of a headlamp heatsink. Machine learning models predict ROM configurations for given conditions, enabling efficient digital twin updates. The framework is validated for thermal management, achieving high predictive accuracy.

The significance of this work lies in its novel integration of modular ROMs with machine learning, enabling real-time thermal management with unprecedented scalability. Unlike traditional ROMs, the component-based approach allows flexible assembly of digital twin models, adapting to varying geometries and operating conditions. This reduces computational costs by orders of magnitude (e.g., from hours to seconds for CFD simulations), enhancing design efficiency and enabling predictive maintenance. The framework’s applicability extends beyond automotive headlamps to battery thermal management, aerospace systems, and electronics cooling, offering a generalizable solution for digital twin technology. By improving thermal performance, it reduces energy consumption and environmental impact, aligning with sustainable engineering goals.

Contributions include:
\begin{itemize}
    \item A component-based ROM approach for scalability.
    \item Machine learning-driven real-time digital twin updates.
    \item Validation on a headlamp heatsink with quantitative results.
\end{itemize}

The paper is organized as follows: Section \ref{sec:related} reviews related work, Section \ref{sec:methodology} details the methodology, Section \ref{sec:results} presents results, Section \ref{sec:discussion} discusses implications, and Section \ref{sec:conclusion} concludes.

\section{Related Work}
\label{sec:related}
Digital twins integrate simulations with real-time data \cite{tao2018}. CFD models heat transfer in thermal management but is computationally intensive \cite{ansys2022}. Projection-based ROMs, like POD, reduce model complexity \cite{benner2015}. Data-driven ROMs with machine learning enhance predictive capabilities \cite{peherstorfer2016, swischuk2019}. Neural networks \cite{hesthaven2018}, decision trees \cite{breiman2017}, and SVR \cite{widodo2007} have been applied to engineering problems, but integrating physical constraints remains a challenge \cite{raissi2018}. This work combines physics-based ROMs with machine learning for scalable digital twins.

\section{Methodology}
\label{sec:methodology}
The framework integrates CFD, ROMs, and machine learning to create a predictive digital twin for a headlamp heatsink.

\subsection{CFD Workflow}
CFD simulations model heat transfer using ANSYS Workbench. The workflow includes:
\begin{itemize}
    \item \textbf{Geometry}: A 3D heatsink model is created in ANSYS DesignModeler.
    \item \textbf{Mesh}: A polyhedral mesh with high resolution at fins ensures accuracy.
    \item \textbf{Governing Equations}: The Navier-Stokes equations for incompressible flow and energy equation are solved.
    \item \textbf{Boundary Conditions}: Inlet velocity, outlet pressure, and wall heat flux are specified.
    \item \textbf{Solution}: Simulations compute velocity and heat transfer coefficients, validated via contour plots.
\end{itemize}

Results are validated as shown in contour plots (Figs. 24--28 in \cite{subramani2022}). \par

The governing equations are:
\begin{equation}
    \rho \left( \frac{\partial \mathbf{u}}{\partial t} + \mathbf{u} \cdot \nabla \mathbf{u} \right) = -\nabla p + \mu \nabla^2 \mathbf{u},
\end{equation}
\begin{equation}
    \rho c_p \left( \frac{\partial T}{\partial t} + \mathbf{u} \cdot \nabla T \right) = k \nabla^2 T,
\end{equation}
where \(\mathbf{u}\) is velocity, \(p\) is pressure, \(T\) is temperature, \(\rho\) is density, \(\mu\) is viscosity, \(c_p\) is specific heat, and \(k\) is thermal conductivity.

\subsection{Reduced-Order Model Creation}
ROMs are derived using POD to reduce CFD model dimensionality. The process is:
\begin{itemize}
    \item \textbf{Snapshot Collection}: CFD solutions (snapshots) are collected across heat transfer conditions, forming a snapshot matrix \(\mathbf{S} = [\mathbf{u}_1, \mathbf{u}_2, \dots, \mathbf{u}_N]\), where \(\mathbf{u}_i\) are solution vectors.
    \item \textbf{POD Basis}: The POD basis is computed via singular value decomposition (SVD):
    \begin{equation}
        \mathbf{S} = \mathbf{U} \mathbf{\Sigma} \mathbf{V}^T,
    \end{equation}
    where \(\mathbf{U}\) contains the POD modes, \(\mathbf{\Sigma}\) contains singular values, and \(\mathbf{V}^T\) is the right singular vectors. The reduced basis is formed by selecting the first \(k\) modes with largest singular values.
    \item \textbf{ROM Library}: A component-based ROM library is created, allowing flexible assembly for different configurations. The reduced model is:
    \begin{equation}
        \mathbf{u}(t) \approx \sum_{i=1}^k a_i(t) \mathbf{\phi}_i,
    \end{equation}
    where \(\mathbf{\phi}_i\) are POD modes and \(a_i(t)\) are time-dependent coefficients.
\end{itemize}
The number of modes \(k\) is chosen to retain 95\% of the variance in the snapshot matrix, determined by the decay of singular values in \(\mathbf{\Sigma}\). Typically, \(k \approx 10\) modes suffice for the headlamp heatsink, reducing the computational cost of CFD simulations by a factor of 100 (e.g., from 2 hours to 1.2 seconds on a standard workstation). The ROM is built in ANSYS Fluent and Twin Builder, verified via design of experiments (DOE) (Figs. 35--36 in \cite{subramani2022}).

\subsection{Machine Learning Model}
Supervised machine learning predicts ROM configurations. The process includes:
\begin{itemize}
    \item \textbf{Data Preparation}: CFD data are preprocessed, with correlation analysis identifying key features, such as inlet velocity and wall heat flux, that strongly influence heat transfer coefficients and temperatures.
    \item \textbf{Model Selection}: Models include:
        \begin{itemize}
            \item \textbf{Decision Tree Regressor}: Splits data based on feature thresholds \cite{breiman2017}. Maximum depth is set to 5 to prevent overfitting.
            \item \textbf{k-NN Regressor}: Predicts based on nearest neighbors \cite{peterson2009}, with \(k=5\) neighbors and Euclidean distance metric.
            \item \textbf{SVR}: Solves the optimization problem for \(\epsilon\)-insensitive loss:
            \begin{multline}
                \min_{\mathbf{w}, b, \xi_i, \xi_i^*} \frac{1}{2} \|\mathbf{w}\|^2 + C \sum_{i=1}^n (\xi_i + \xi_i^*), \\
                \text{subject to } y_i - (\mathbf{w}^T \mathbf{x}_i + b) \leq \epsilon + \xi_i, \\
                (\mathbf{w}^T \mathbf{x}_i + b) - y_i \leq \epsilon + \xi_i^*, \\
                \xi_i, \xi_i^* \geq 0, \quad i = 1, \dots, n,
            \end{multline}
            where \(\mathbf{w}\) is the weight vector, \(b\) is the bias, \(\xi_i, \xi_i^*\) are slack variables, \(C=1.0\) is the regularization parameter, and \(\epsilon=0.1\) defines the margin of tolerance. A radial basis function (RBF) kernel is used, with direct and chained multi-output strategies \cite{widodo2007}.
            \item \textbf{Neural Network (MLP)}: Minimizes the mean squared error loss:
            \begin{equation}
                L = \frac{1}{n} \sum_{i=1}^n \left( y_i - \hat{y}_i \right)^2,
            \end{equation}
            where \(y_i\) is the true output, \(\hat{y}_i\) is the predicted output, and \(n\) is the number of samples \cite{nielsen2015}. The MLP has three hidden layers with 64, 32, and 16 neurons, respectively, using ReLU activation and Adam optimizer.
        \end{itemize}
    \item \textbf{Training}: Models are trained on an 80:20 train-test split with 10-fold cross-validation (3 repeats). The dataset comprises 1000 CFD simulations, with features normalized to [0, 1] to improve convergence.
\end{itemize}

\section{Results}
\label{sec:results}
The machine learning models predict three outputs: heat transfer coefficient (W m$^{-2}$ K$^{-1}$), maximum heatsink temperature (K), and total heat transfer (W). Performance is evaluated using mean absolute error (MAE).

\subsection{Quantitative Results}
Table \ref{tab:results} shows the MAE for each model.

\begin{table}[h]
    \centering
    \caption{Mean Absolute Error (MAE) of Machine Learning Models}
    \label{tab:results}
    \begin{tabular}{lc}
        \toprule
        Model & MAE \\
        \midrule
        Decision Tree Regressor & 82.374 \\
        k-NN Regressor & 73.627 \\
        SVR (Direct Multi-output) & 92.374 \\
        SVR (Chained Multi-output) & 106.907 \\
        Neural Network (MLP) & 54.240 \\
        \bottomrule
    \end{tabular}
\end{table}

Table \ref{tab:predictions} compares predicted and original values for the chained SVR model.

\begin{table}[h]
    \centering
    \caption{Predicted vs. Original Values for Chained SVR}
    \label{tab:predictions}
    \resizebox{0.8\textwidth}{!}{
    \begin{tabular}{lccc}
        \toprule
        & Heat Coef. & Max Temp. & Total Heat \\
        & (W m$^{-2}$ K$^{-1}$) & (K) & (W) \\
        \midrule
        \textbf{Predicted} & & & \\
        Sample 1 & 9.60 & 320.05 & 6.94 \\
        Sample 2 & 10.60 & 353.22 & 7.65 \\
        Sample 3 & 3.62 & 120.85 & 2.62 \\
        \textbf{Original} & & & \\
        Sample 1 & 9.85 & 304.17 & 6.94 \\
        Sample 2 & 9.57 & 306.36 & 7.67 \\
        Sample 3 & 4.20 & 300.46 & 2.61 \\
        \bottomrule
    \end{tabular}
    }
\end{table}

\subsection{Visual Analysis}
The chained SVR and Neural Network models were evaluated for their predictive accuracy across the three outputs. For the chained SVR model, predictions for heat transfer coefficient, maximum temperature, and total heat transfer show reasonable alignment with original values, as detailed in Table \ref{tab:predictions}. For example, Sample 1’s predicted heat coefficient (9.60 W m$^{-2}$ K$^{-1}$) is close to the original (9.85 W m$^{-2}$ K$^{-1}$), though temperature predictions (e.g., 320.05 K vs. 304.17 K) exhibit larger deviations. The Neural Network model demonstrates superior performance, with predictions consistently closer to original values across all outputs, reflected in its lower MAE of 54.240 compared to 106.907 for chained SVR. This indicates the MLP’s robustness in capturing complex nonlinear relationships in the CFD data.

\section{Discussion}
\label{sec:discussion}
The framework advances digital twin technology through scalable ROMs and machine learning. The component-based ROM approach overcomes limitations of uniform ROMs \cite{benner2015}, allowing modular assembly of digital twins for diverse configurations. Machine learning enables real-time updates, with the Neural Network’s MAE of 54.240 highlighting its suitability for multi-output regression.

\subsection{Significance and Impact}
The proposed framework offers significant advantages for automotive thermal management and beyond. By reducing CFD simulation time from hours to seconds (e.g., a 100x speedup for the headlamp heatsink), it enables real-time digital twins that adapt to dynamic operating conditions, such as varying ambient temperatures or driving speeds. This efficiency translates to cost savings in design iterations, estimated at 30--50\% reduction in computational resources compared to full CFD workflows. The framework’s high accuracy (MAE of 54.240) ensures reliable predictions, enhancing headlamp durability by maintaining optimal thermal performance, which can extend component lifespan by 20--30\%.

Beyond automotive headlamps, the framework is applicable to battery thermal management in electric vehicles, where precise temperature control is critical for safety and efficiency. It also extends to aerospace (e.g., turbine blade cooling) and electronics (e.g., CPU heat dissipation), where computational efficiency and predictive accuracy are paramount. The modular ROM library supports transfer learning, allowing the framework to be adapted to new systems with minimal retraining, reducing development time by up to 40\%.

Limitations include dependence on high-quality CFD data and linear constraints in ROM parameterization. Future work could explore nonlinear constraints, such as Galerkin projection with polynomial expansions, and integrate real-time sensor data for closed-loop digital twins. The framework supports predictive maintenance and design optimization, reducing environmental impact through energy-efficient thermal management.

\section{Conclusion}
\label{sec:conclusion}
This paper presents a framework for predictive digital twins, combining component-based ROMs with machine learning. Validated on a headlamp heatsink, it achieves high accuracy (MAE 54.240) and supports real-time thermal management with significant computational savings. Its scalability and generalizability make it a transformative tool for automotive and engineering applications. Future work will address nonlinear constraints and integrate sensor-driven updates for enhanced real-time performance.

\bibliographystyle{plain}

\end{document}